\documentclass{article}

\usepackage[final]{neurips_2021}

\usepackage[utf8]{inputenc} 
\usepackage[T1]{fontenc}    
\usepackage{hyperref}       
\usepackage{url}            
\usepackage{booktabs}       
\usepackage{amsfonts}       
\usepackage{nicefrac}       
\usepackage{microtype}      
\usepackage{xcolor}         
\usepackage{graphics}
\usepackage{graphicx}
\usepackage{multirow}

\title{Evaluation of mathematical questioning strategies using data collected through weak supervision}

\author{%
  Debajyoti Datta\\
  SEAS, University of Virginia\\
  Charlottesville, VA 22903 \\
  USA \\
  \texttt{dd3ar@virginia.edu} \\
  \And Maria Phillips \\
   SEAS, University of Virginia\\
   Charlottesville, VA 22903 \\
   USA \\
   \And James P Bywater \\
   James Madison University\\
   Charlottesville, VA 22903 \\
   USA \\
   \And Jennifer Chiu \\
   School of Education and Human Development\\
   University of Virginia\\
   Charlottesville, VA 22903 \\
   USA \\
    \And Ginger S. Watson \\
   School of Education and Human Development\\
   University of Virginia\\
   Charlottesville, VA 22903 \\
   USA \\
    \And Laura E. Barnes \\
  SEAS, University of Virginia\\
  Charlottesville, VA 22903 \\
  USA \\
    \And Donald E Brown \\
  SEAS, University of Virginia\\
  Charlottesville, VA 22903 \\
  USA \\
}

\begin{document}

\maketitle

\begin{abstract}
A large body of research demonstrates how teachers' questioning strategies can improve student learning outcomes. However, developing new scenarios is challenging because of the lack of training data for a specific scenario and the costs associated with labeling. This paper presents a high-fidelity, AI-based classroom simulator to help teachers rehearse research-based mathematical questioning skills. Using a human-in-the-loop approach, we collected a high-quality training dataset for a mathematical questioning scenario. Using recent advances in uncertainty quantification, we evaluated our conversational agent for usability and analyzed the practicality of incorporating a human-in-the-loop approach for data collection and system evaluation for a mathematical questioning scenario.
\end{abstract}

\section{Introduction}

Real-world applications of deep learning models require hand-labeled training data specific to the domain. While the natural language processing community has divested a significant effort in collecting datasets for popular tasks like sentiment analysis or textual entailment, domain-specific datasets like teacher questioning often lack high-quality labeled datasets. In this work, we use weak supervision to collect a teacher questioning dataset for a mathematical training scenario and use an expert-in-the-loop system to deploy and collect more training data in the process.
\subsection{Scenario}

\begin{figure}[t]
\centering
\includegraphics[width=1\textwidth]{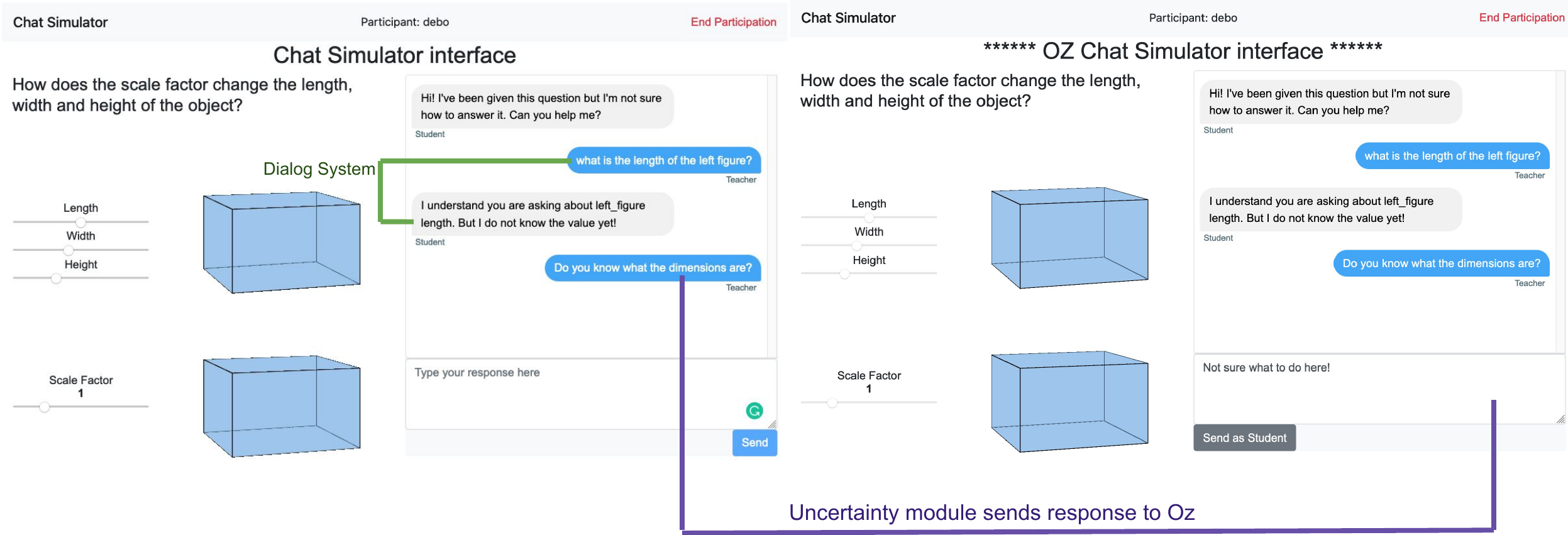}
\includegraphics[width=0.5\textwidth]{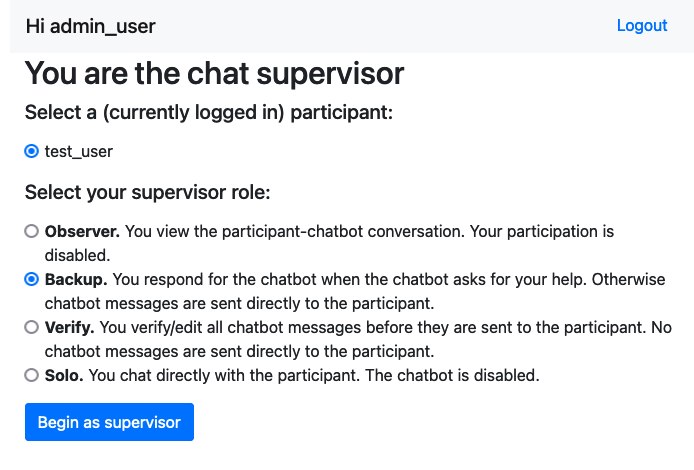} 
\caption{\textbf{The expert user has access to the Supervisor Interface to the right. The dialogue system responds when it is certain about the dialogue acts and entities. When the uncertainty thresholds are met, it sends the prompt to the expert user. The expert user can then type in the response as a student, and it will appear on the prompt (left). This prevents conversations from breaking because of the failure of ML pipelines and dialogue system components.}}
\label{fig_scale_factor}
\end{figure}

The conversational agent (CA) is a student who is learning the concept of scale factor, and the user of the system is a pre-service teacher in training. In traditional classroom scenarios, teachers often use visual aids to explain various mathematical concepts. We developed an HTML5 based interactive widget that the pre-service teacher can use to engage in mathematical discussion with the virtual students (Figure \ref{fig_scale_factor}). One main goal of this dialogue system is to help pre-service teachers develop mathematical question-asking strategies through rehearsal with a CA. To evaluate pre-service teachers' dialogue and provide feedback on performance, the CA uses an adapted rubric of the Instructional Quality Assessment(IQA) focused on instructional questions \citep{boston2012assessing, boston2018instructional}. In our adapted IQA rubric, questions are categorized into probing or exploring, factual or recall, expository or cueing, and other (Table \ref{dialogueacts}). Table \ref{perform_comparison} provides specific details of the data collected by two annotators using weak supervision for the dialogue system components.

\section{Challenges in data collection and evaluation}

Effective questioning strategies from instructors in the classroom can improve student learning outcomes \citep{kilgo2015link, ellis1993teacher, cotton1988classroom, wilen1986effective}, but practicing such questions systematically and deliberately during teacher education programs and opportunities can be difficult. For example, in a traditional conversational agent (CA) for flight booking, the CA might only have to answer very domain-specific questions like ``Where are you flying to?'', but in a mathematics classrooms questions are generally more open-ended like ``How did you get that answer?''. Similar probing questions, as in Table \ref{dialogueacts} are often challenging to evaluate because there is a subjective component that depends significantly on the context and the previous utterances. In education, CAs used with students directly have resulted in a variety of learning outcomes, \citep{d2014confusion} including improved learning of mathematics concepts like Mathbot \citep{graesser2014learning, grossman2019mathbot} and writing skills \citep{li2021impact}. However, very few of these systems focus on pre-service teachers, and none to our knowledge incorporate a human-centered approach for facilitating a task-specific conversation.


\noindent\textbf{Weak supervision for data labeling}: Weak supervision is a machine learning paradigm that trains models by incorporating noisy labeled data. These noisy labels are either crowd-sourced \citep{dawid1979maximum} or created through machine learning models and labeling functions \citep{ratner2016data}. After using weak supervision for the initial data labeling stages, we use model-assisted labeling \citep{LabelStudio} to speed up the annotators' labeling process. Model-assisted labeling significantly benefits the annotation pipeline since expert annotators take 70\% less time to label data. One advantage of using weak supervision with expert labelers is that they can effectively develop nuanced labeling functions with higher coverage which speeds up annotation in the model-assisted labeling stage. It is worth noting that when dataset sizes are large weak supervision achieves accuracies similar to supervised labeled datasets \citep{ratner2016data}. Since the dataset size is small in our context, it is used as part of the data labeling pipeline and not by itself. 

\section{Uncertainty based dialogue system}

\noindent\textbf{Dialogue Act Module}: The intent classification stage quantifies the uncertainty between the different categories of dialogue acts. For uncertainty measurement, we use Active Dropout \citep{gal2016dropout} since it is easy to adapt to existing models with minimal change. The dialogue acts are shown in Table \ref{dialogueacts}. Based on the module-specific threshold of the uncertainty as shown in Figure \ref{fig_scale_factor}, we determine if the response should be sent to the supervisor. This can be a complex query, a failure to match an intent or an out-of-domain query like ``Have you seen the recent Batman movie?''. If the uncertainty module gets triggered by the threshold, the supervisor intervenes and states the system's limitations or answers the question if it was because of a failure of one of the components. 
The average \textbf{F1 score was 0.71} for the dataset collected through weak supervision. The model used was DistilBERT \citep{sanh2019distilbert}.  

\noindent\textbf{Entity Recognition Module}: The entity recognition module was a NER (Named Entity Recognition Module) module trained with Spacy \citep{spacy}. Spacy's rule-based pattern `matchers' can infer complex patterns from the text by using linguistic properties of words (like part-of-speech tags) and regular expressions. The rule-based matcher was then used to create the dataset  which was subsequently trained on a logistic regression classifier similar to the approach by \citep{nadav2016}. We used pre-trained BERT sentence representations to train the logistic regression classifier (\textbf{precision: 0.84, recall: 0.82, f1: 0.83}). The entity extraction task was framed as shown in Table \ref{entity_extraction} (In Appendix). 

\noindent\textbf{Turn-Taking Module}: Turn coherence and turn-based uncertainty are critical in conversational agents since conversations can derail with an out-of-context response \citep{lin2019moel} or an inappropriate response \citep{young2018augmenting}. To find if users are asking semantically equivalent questions in consecutive turns, we match the semantic similarity of the previous user question and the new user question using Sentence-BERT \citep{reimers-2019-sentence-bert}. The cosine similarity between the embeddings gives a score we use to compute if the utterance is very similar to the previous utterance.

\begin{table*}[]
\scalebox{0.6}{
\begin{tabular}{llll}
\hline
\textbf{Question label}                                                                                       & \textbf{Description}                                                                                                                                                                                                                                                                 & \multicolumn{1}{l}{\textbf{Examples}}                                                                                                                                                                                & \textbf{\begin{tabular}[c]{@{}l@{}}Weak Supervision \\ Pattern Example\end{tabular}}                                                                                                   \\ \hline
\begin{tabular}[c]{@{}l@{}}Probing or \\ exploring mathematical\\  meanings and \\ relationships\end{tabular} & \begin{tabular}[c]{@{}l@{}}Question clarifies student thinking, enables \\ students to elaborate their own thinking for \\ their own benefit and the class; Points to \\ underlying mathematical relationships\\  and meanings; Makes links among \\ mathematical ideas\end{tabular} & \multicolumn{1}{l}{\begin{tabular}[c]{@{}l@{}}How did you get that answer?\\ Explain to me how you got that expression?\\ What does n represent in terms of the diagram?\\ Why is it staying the same?\end{tabular}} & \begin{tabular}[c]{@{}l@{}}Includes words/phrases ``how'', \\ ``how did you'', ``why did \\ you'',  ``what is staying the same'', \\ ``explain to me'', ``how could you''\end{tabular} \\ \hline
Factual or recall                                                                                             & \begin{tabular}[c]{@{}l@{}}Elicits a mathematical fact; Requires \\ a single response answer; Requires \\ the recall of a memorized fact or \\ procedure, can be a yes/no answer but \\ for a specific mathematical question\end{tabular}                                            & \begin{tabular}[c]{@{}l@{}}What is 3x5?\\ Does this picture show ½ or ¼?\\ What do you subtract first?\end{tabular}                                                                                                   & \begin{tabular}[c]{@{}l@{}}Includes words/phrases: \\ ``what is this'', ``what is next'', \\ ``what is {[}fact{]}'', \\ ``what would you do next''\end{tabular}                        \\ \hline
Expository or cueing                                                                                          & \begin{tabular}[c]{@{}l@{}}Provides mathematical cueing or \\ mathematical information to students, \\ tells them to look at specific information\\  without engaging students’ ideas\end{tabular}                                                                                   & \begin{tabular}[c]{@{}l@{}}Rhetorical questions (“The answer is three, right?”)\\ Clarifying statements “Between the 2?”\\ Look at this diagram\end{tabular}                                                          & \begin{tabular}[c]{@{}l@{}}Includes words/phrases: \\ ``right?'', ``and then you'', \\ ``then I'', ``this is''\end{tabular}                                                            \\ \hline
Other                                                                                                         & \begin{tabular}[c]{@{}l@{}}Non-academic behavioral talk; General \\ classroom management; everything else.\end{tabular}                                                                                                                                                              & \begin{tabular}[c]{@{}l@{}}Sit down\\ Close your books\end{tabular}                                                                                                                                                   & \begin{tabular}[c]{@{}l@{}}Semantic similarity with a \\ collection of examples labeled as\\ ``Others''\end{tabular}                                                                   \\ \hline
\end{tabular}}
\caption{\label{dialogueacts}
Weak supervision provides an effective way to generate pre-filled labels that can be used with model-assisted labeling to increase annotators' speed.
}
\end{table*}

\begin{table*}[]
\scalebox{0.9}{
\begin{tabular}{ccccccccc}
\hline
\multicolumn{2}{c}{\multirow{2}{*}{\textbf{Technique}}}                                                                            & \multirow{2}{*}{\textbf{Annotator}} & \multirow{2}{*}{\textbf{n}} & \multirow{2}{*}{\textbf{\begin{tabular}[c]{@{}c@{}}Annotator \\ Accuracy\end{tabular}}} & \multirow{2}{*}{\textbf{\begin{tabular}[c]{@{}c@{}} Agreement\\ (kappa)\end{tabular}}} & \multirow{2}{*}{\textbf{\begin{tabular}[c]{@{}c@{}}Model-\\ Assisted \\ Agreement\end{tabular}}} & \multicolumn{2}{c}{\textbf{Time}}                                                                      \\ \cline{8-9} 
\multicolumn{2}{c}{}                                                                                                               &                                     &                             &                                                                                         &                                                                                                            &                                                                                                  & \textbf{\begin{tabular}[c]{@{}c@{}}M(SD)\\ (seconds)\end{tabular}} & \textit{\textbf{p-value}}         \\ \hline
\multirow{3}{*}{A} & \multirow{3}{*}{\begin{tabular}[c]{@{}c@{}}Classical \\ Labeling\end{tabular}}                                & Both                                & 1730                        & 0.82                                                                                    & \multirow{3}{*}{0.52}                                                                                      & -                                                                                                & 15.2(42.1)                                                         & \multirow{6}{*}{\textless{}0.001} \\ \cline{3-5} \cline{7-8}
                   &                                                                                                               & A1                                  & 864                         & 0.89                                                                                    &                                                                                                            & -                                                                                                & 13.2(37.8)                                                         &                                   \\ \cline{3-5} \cline{7-8}
                   &                                                                                                               & A2                                  & 866                         & 0.74                                                                                    &                                                                                                            & -                                                                                                & 17.3(45.9)                                                         &                                   \\ \cline{1-8}
\multirow{3}{*}{B} & \multirow{3}{*}{\begin{tabular}[c]{@{}c@{}}WS - MAL\end{tabular}} & Both                                & 3983                        & 0.84                                                                                    & \multirow{3}{*}{0.61}                                                                                      & 0.70                                                                                             & 10.4(32.7)                                                         &                                   \\ \cline{3-5} \cline{7-8}
                   &                                                                                                               & B1                                  & 1994                        & 0.89                                                                                    &                                                                                                            & 0.80                                                                                             & 7.1(18.3)                                                          &                                   \\ \cline{3-5} \cline{7-8}
                   &                                                                                                               & B2                                  & 1989                        & 0.79                                                                                    &                                                                                                            & 0.60                                                                                             & 13.6(42.3)                                                         &                                   \\ \cline{1-8}
\end{tabular}
}
\caption{\label{perform_comparison}
Performance comparison of annotators between traditional supervised labeling and weak supervision + model-assisted labeling approach. The gold accuracy table refers to anonymized annotators accuracy compared to gold labels generated by expert teachers with significant experience in evaluating IQA metrics.
}

\end{table*}


\section{Experimental Setting}

\begin{table}[]
\caption{\label{ws_data_label} \textbf{Certain utterances were always sent to the supervisor. However for most cases, the NLP components could correctly classify the utterances.}}
\scalebox{0.7}{
\begin{tabular}{llll}
\hline
\textbf{Collaboration Approach}   & \textbf{Definitions and Examples}                                                                                                                                                                                                                                                                                                                                                                                                                      & \textbf{Number of Occurances}                                                 & \textbf{Varied Based On Turn}                                                      \\ \hline
\textbf{Always Supervisor}        & \begin{tabular}[c]{@{}l@{}}Chit-Chat type questions\\ "What is your name?"\\ \\ Detailed explanation questions\\ "Can you explain why the volume is 25?"\\ \\ Out of domain questions\\ "Can you calculate it for a sphere?"\\ \\ Overly complex descriptions\\ "If all the dimensions of the left box is 5 and you \\ change the scale factor by 2, and all other \\ dimensions remain unchanged what is the volume \\ of the right box?\end{tabular} & \begin{tabular}[c]{@{}l@{}}6\\ \\ \\ 8\\ \\ \\ \\ 3\\ \\ \\ \\ 5\end{tabular} & \begin{tabular}[c]{@{}l@{}}No\\ \\ \\ No\\ \\ \\ \\ No\\ \\ \\ \\ Yes\end{tabular} \\ \hline
\textbf{Always AI}                & \begin{tabular}[c]{@{}l@{}}Question about dimensions\\ "What is the right figure volume?"\\ "What is the scale factor?"\\ \\ Greetings\\ "Hi, How are you?"\\ \\ Simple Explanations\\ "Do you know how to calculate the volume?"\end{tabular}                                                                                                                                                                                                         & \begin{tabular}[c]{@{}l@{}}14\\ \\ \\ \\ 5\\ \\ \\ 16\end{tabular}            & \begin{tabular}[c]{@{}l@{}}No\\ \\ \\ \\ No\\ \\ \\ No\end{tabular}                \\ \hline
\textbf{Mix of AI and Supervisor} & \begin{tabular}[c]{@{}l@{}}Definitions\\ "How did you calculate the scale factor?"\\ "Would the volume have changed if the scale factor was 1?"\\ \\ Acknowledgment\\ "Great job!", "Oh nice!",  "Very good!"\\ \\ Task Status\\ "So do you think you understand the task at hand?"\end{tabular}                                                                                                                                                       & \begin{tabular}[c]{@{}l@{}}7\\ \\ \\ \\ 8\\ \\ \\ 2\end{tabular}              & \begin{tabular}[c]{@{}l@{}}No\\ \\ \\ \\ No\\ \\ \\ No\end{tabular}                \\ \hline
\end{tabular}
}
\end{table}

The evaluation included eight users with teaching experience (ages 24-60), each conversing with the system twice. The analysis of the user utterances are shown in Table \ref{ws_data_label}. Certain utterances (``Always AI'' and ``Mix of AI and supervisor'') benefit from weak-supervision-based data labeling approaches. Complex queries that always go to the supervisor can be used for future data labeling scenarios.

\section{Conclusion and Future Work}
Our goal in this paper was to build a conversational agent using weak supervision for a mathematical training scenario. Our analysis showed that conversational agents could be built for mathematical scenarios that do not have large datasets by combining weak supervision and accounting for model failures. Uncertainty quantification and training with noisy labels like weak supervision is an active area of research, and more robust training paradigms will help build newer scenarios with minimal data. This is especially useful in educational research where data collection predominantly relies on video and audio transcriptions of classroom recordings. In the future, we plan to incorporate other modalities like speech into the system to understand and evaluate limitations of uncertainty in other modalities other than text.
\bibliographystyle{plainnat}
\bibliography{sample}

\appendix

\section{Appendix}

\begin{table}[htb]
\caption{\textbf{Entity extraction task for uncertainty modeling}}
\begin{tabular}{llll}
\hline
\textbf{Text}                                     & \textbf{Entities} & \textbf{Relation}    & \textbf{Label} \\ \hline
The length of the object is 5, what is the width? & length, 5, width  & (length, 5)          & True           \\
What is the scale factor?                         & scale factor      & (scale factor, \_\_) & False          \\
No, the length is not 5, the width is.            & length, 5, width  & (width, 5)           & True           \\
No, the length is not 5, the width is.            & length, 5, width  & (length, 5)          & False          \\ \hline
\label{entity_extraction}
\end{tabular}
\end{table}

\end{document}